\documentclass[11pt, a4paper, nocopyright]{deepmind}

\usepackage[authoryear, sort&compress, round]{natbib}

\usepackage{placeins}
\usepackage[toc,page]{appendix}

\usepackage{nicefrac}       

\usepackage{boldline}       

\usepackage[font=small,labelfont=bf]{subcaption}
\usepackage[font=small]{caption}
\usepackage{verbatim}
\usepackage{xpatch}
\usepackage{setspace}
\usepackage{adjustbox}
\usepackage{amsmath, latexsym}
\usepackage{amsfonts}
\usepackage{amsthm}
\usepackage{xspace}
\usepackage{mathtools}

\usepackage{xcolor}
\definecolor{numbersColor}{rgb}{0.5, 0.5, 0.5}
\definecolor{commentsColor}{rgb}{0.297495, 0.597587, 0.297464}
\definecolor{keywordsColor}{rgb}{0.000000, 0.000000, 0.635294}
\definecolor{stringColor}{rgb}{0.558215, 0.000000, 0.135316}
\usepackage{listings}
\usepackage{multirow}
\usepackage{arydshln}
\usepackage{rotating}
\usepackage{array}
\usepackage{mfirstuc}

\usepackage{amsmath, latexsym}
\usepackage{amsfonts}
\usepackage{mathtools}
\usepackage{amsthm}
\usepackage{dsfont}
\usepackage[dvipsnames]{xcolor}
\usepackage[colorinlistoftodos]{todonotes}
\usepackage{booktabs}
\usepackage{xfrac}
\usepackage{bbm}

\usepackage{xparse}
\usepackage{lipsum}
\usepackage{changepage}
\usepackage{enumitem}

\newcommand{\squishlist}{
   \begin{list}{$\bullet$}
    { \setlength{\itemsep}{0pt}      \setlength{\parsep}{3pt}
      \setlength{\topsep}{3pt}       \setlength{\partopsep}{0pt}
      \setlength{\leftmargin}{1.5em} \setlength{\labelwidth}{1em}
      \setlength{\labelsep}{0.5em} } }

\newcommand{\squishlisttwo}{
   \begin{list}{$\bullet$}
    { \setlength{\itemsep}{0pt}    \setlength{\parsep}{0pt}
      \setlength{\topsep}{0pt}     \setlength{\partopsep}{0pt}
      \setlength{\leftmargin}{2em} \setlength{\labelwidth}{1.5em}
      \setlength{\labelsep}{0.5em} } }

\newcommand{\squishend}{
    \end{list}  }

\newcolumntype{C}[1]{>{\centering\arraybackslash}m{#1}}
\newcolumntype{R}[1]{>{\raggedleft\arraybackslash}m{#1}}

\newcommand{\two}{(2)}

\DeclareMathAlphabet{\mathpzc}{OT1}{pzc}{m}{n}

\usepackage{makecell}
\usepackage{fancyvrb}

\RequirePackage{algorithm}
\RequirePackage{algorithmic}
\renewcommand{\algorithmiccomment}[1]{\hfill$\blacktriangleright$ #1}
\renewcommand{\algorithmicrequire}{\textbf{Input:}}
\renewcommand{\algorithmicensure}{\textbf{Output:}}

\correspondingauthor{nwatters@mit.edu}

\title{Modular Object-Oriented Games: A Task Framework for Reinforcement Learning, Psychology, and Neuroscience}

\author[1]{Nicholas Watters}
\author[1,2]{Joshua Tenenbaum}
\author[1,3]{Mehrdad Jazayeri}

\affil[1]{Department of Brain and Cognitive Sciences, Massachusetts Institute of Technology}
\affil[2]{Center for Brains, Minds and Machines, MIT}
\affil[3]{McGovern Institute of Brain Research, MIT}

\renewcommand{\today}{2021-02-17}

\begin{abstract}

In recent years, trends towards studying simulated games have gained momentum in the fields of artificial intelligence, cognitive science, psychology, and neuroscience.
The intersections of these fields have also grown recently, as researchers increasing study such games using both artificial agents and human or animal subjects.
However, implementing games can be a time-consuming endeavor and may require a researcher to grapple with complex codebases that are not easily customized.
Furthermore, interdisciplinary researchers studying some combination of artificial intelligence, human psychology, and animal neurophysiology face additional challenges, because existing platforms are designed for only one of these domains.
Here we introduce \href{https://jazlab.github.io/moog.github.io/moog/index.html}{Modular Object-Oriented Games}, a Python task framework that is lightweight, flexible, customizable, and designed for use by machine learning, psychology, and neurophysiology researchers.
\end{abstract}

\begin{document}
\maketitle
\balance

\vskip 0.3in

\vspace*{-10pt}
\section{Introduction}\label{S:intro}

In recent years, trends towards studying object-based games have gained momentum in the fields of artificial intelligence, cognitive science, psychology, and neuroscience.
In artificial intelligence, interactive physical games are now a common testbed for reinforcement learning \citep{mnih2013playing, sutton2018reinforcement, leike2017ai, Francois_Lavet_2018} and object representations are of particular interest for sample efficient and generalizable AI \citep{battaglia2018relational, van2019perspective, greff2020binding}.
In cognitive science and psychology, object-based games are used to study a variety of cognitive capacities, such as planning, intuitive physics, and intuitive psychology \citep{ullman2017mind, Chabris2017}.
Developmental psychologists also use object-based visual stimuli to probe questions about object-oriented reasoning in infants and young animals \citep{spelke2007core, wood2020reverse}.
In neuroscience, object-based computer games have recently been used to study decision-making and physical reasoning in both human and non-human primates \citep{fischer2016functional, mcdonald2019bayesian, yoo2020neural, Rajalingham2021}.

Furthermore, a growing number of researchers are studying tasks using a combination of approaches from these fields.
Comparing artificial agents with humans or animals performing the same tasks can help constrain models of human/animal behavior, generate hypotheses for neural mechanisms, and may ultimately facilitate building more intelligent artificial agents \citep{lake2017building, willke2019comparison, hassabis2017neuroscience}.

However, building a task that can be played by AI agents, humans, and animals is a time-consuming undertaking because existing platforms are typically designed for only one of these purposes.
Professional game engines are designed for human play and are often complex libraries that are difficult to customize for training AI agents and animals.
Reinforcement learning platforms are designed for AI agents but are often too slow or inflexible for neuroscience work.
Existing psychology and neurophysiology platforms are too limited to easily support complex interactive games.

In this work we offer a game engine that is highly customizable and designed to support tasks that can be played by AI agents, humans, and animals.

\section{Summary}\label{S:env_tasks}

The \href{https://jazlab.github.io/moog.github.io/index.html}{Modular Object-Oriented Games} library is a general-purpose Python-based platform for interactive games.
It aims to satisfy the following criteria:
\begin{itemize}
    \item
    Usable for reinforcement learning, psychology, and neurophysiology.
    MOOG supports \href{https://github.com/deepmind/dm_env}{DeepMind dm\_env} and \href{https://gym.openai.com/}{OpenAI Gym} \citep{openai_gym} interfaces for RL agents and an \href{https://mworks.github.io/}{MWorks} interface for psychology and neurophysiology.
    
    \item
    Highly customizable.
    Environment physics, reward structure, agent interface, and more can be customized.
    
    \item
    Easy to rapidly prototype tasks.
    Tasks can be composed in a single short file.
    
    \item
    Light-weight and efficient.
    Most tasks run quickly, almost always faster than 100 frames per second on CPU and often much faster than that.
    
    \item
    Facilitates procedural generation for randomizing task conditions every trial.
  
\end{itemize}

\section{Intended Users}\label{S:users}

MOOG was designed for use by the following kinds of researchers:

\begin{itemize}
    \item
    Machine learning researchers studying reinforcement learning in 2.5-dimensional (2-D with occlusion) physical environments who want to quickly implement tasks in Python.
    \item
    Psychology researchers who want more flexibility than existing psychology platforms afford.
    \item
    Neurophysiology researchers who want to study interactive games yet still need to precisely control stimulus timing.
    \item
    Machine learning researchers studying unsupervised learning, particularly in the video domain.
    MOOG can be used to procedurally generate video datasets with controlled statistics.
\end{itemize}

MOOG may be particularly useful for interdisciplinary researchers studying AI
agents, humans, and animals (or some subset thereof) all playing the same task.

\section{Design}\label{S:design}

\begin{figure}[t!]
  \centering
  \includegraphics[width=1.\linewidth]{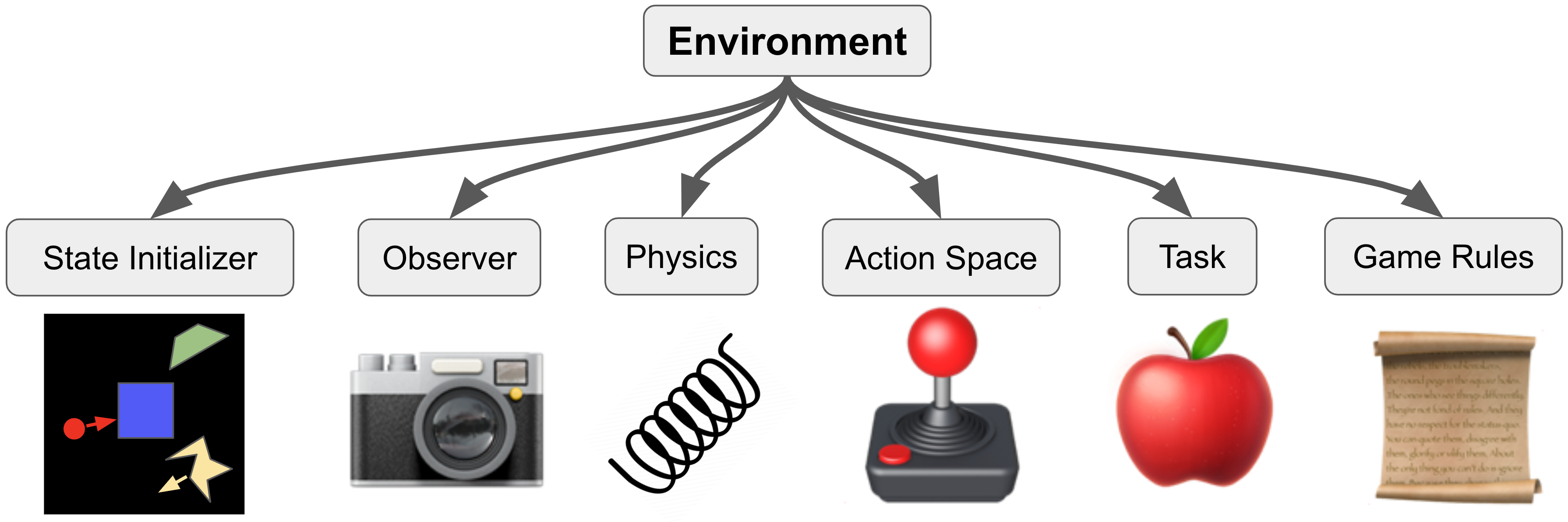}
  \caption{\textbf{Components of a MOOG environment.}
  See main text for details.
  }\label{fig:env_schematic}
\end{figure}

The core philosophy of MOOG is \textbf{"one task, one file."} Namely, a task should be implemented with a single configuration file.
This configuration file is a short ``recipe'' for the task, containing as little substantive code as possible, and should define a set of components to pass to the MOOG environment.
See Figure \ref{fig:env_schematic} for a schematic of these components.

A MOOG environment receives the following components (or callables returning
them) from the configuration file:

\begin{itemize}
    \item
    \textbf{State}.
    The state is a collection of sprites.
    Sprites are polygonal shapes with color and physical attributes (position, velocity, angular velocity, and mass).
    Sprites are 2-dimensional, and the state is 2.5-dimensional with z-ordering for occlusion.
    The initial state can be procedurally generated from a custom distribution at the beginning of each episode.
    The state is structured in terms of layers, which helps hierarchical organization.
    See \href{https://jazlab.github.io/moog.github.io/moog/state_initialization/index.html}{state\_initialization} for procedural generation tools.
    \item
    \textbf{Physics}.
    The physics component is a collection of forces that operate on the sprites.
    There are a variety of forces built into MOOG (collisions, friction, gravity, rigid tethers, ...) and additional custom forces can also be used.
    Forces perturb the velocity and angular velocity of sprites, and the sprite positions and angles are updated with Newton's method.
    See \href{https://jazlab.github.io/moog.github.io/moog/physics/index.html}{physics} for more.
    \item 
    \textbf{Task}.
    The task defines the rewards and specifies when to terminate a trial.
    See \href{https://jazlab.github.io/moog.github.io/moog/tasks/index.html}{tasks} for more.
    \item
    \textbf{Action Space}.
    The action space allows the subject to control the environment.
    Every environment step calls for an action from the subject.
    Action spaces may impart a force on a sprite (like a joystick), move a sprite in a grid (like an arrow-key interface), set the position of a sprite (like a touch-screen), or be customized.
    The action space may also be a composite of constituent action spaces, allowing for multi-agent tasks and multi-controller games.
    See \href{https://jazlab.github.io/moog.github.io/moog/action_spaces/index.html}{action\_spaces} for more.
    \item
    \textbf{Observers}.
    Observers transform the environment state into a observation for the subject/agent playing the task.
    Typically, the observer includes a renderer producing an image.
    However, it is possible to implement a custom observer that exposes any function of the environment state.
    The environment can also have multiple observers.
    See \href{https://jazlab.github.io/moog.github.io/moog/observers/index.html}{observers} for more.
    \item
    \textbf{Game Rules} (optional).
    If provided, the game rules define dynamics or transitions not captured by the physics.
    A variety of game rules are included, including rules to modify sprites when they come in contact, conditionally create new sprites, and control phase structure of trials (e.g. fixation phase to stimulus phase to response phase).
    See \href{https://jazlab.github.io/moog.github.io/moog/game_rules/index.html}{game\_rules} for more.
\end{itemize}

Importantly, all of these components can be fully customized.
If a user would like a physics force, action space, or game rule not provided by MOOG, they can implement a custom one, inheriting from the abstract base class for that component.
This can typically be done with only a few lines of code.

The modularity of MOOG facilitates code re-use across experimental paradigms.
For example, if a user would like to both collect behavior data from humans using a continuous joystick and train RL agents with discrete action spaces on the same task, they can re-use all other components in the task configuration, only changing the action space.

For users interested in doing psychology or neurophysiology, we include an \href{https://github.com/jazlab/moog.github.io/tree/master/mworks}{example} of how to run MOOG through \href{https://mworks.github.io/}{MWorks}, a platform with precise timing control and interfaces for eye trackers, HID devices, electrophysiology software, and more.

\section{Example Tasks}\label{S:examples}

\vspace*{-5pt}

\begin{figure}[t!]
  \centering
  \includegraphics[width=1.\linewidth]{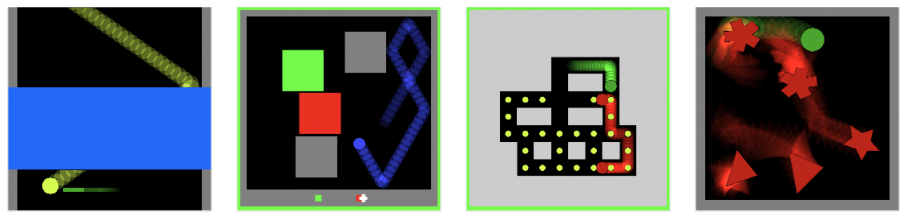}
  \caption{\textbf{Example tasks.}
  Time-lapse images of four example tasks.
  Left-to-right:
  (i) Pong - The subject aims to catch the yellow ball with the green paddle,
  (ii) Red-Green - The subject tries to predict whether the blue ball with contact the red square or the green square,
  (iii) Pac-Man - The subject moves the green agent to catch yellow pellets while avoiding the red ghosts,
  (iv) Collisions - the green agent avoids touching the bouncing polygons.
  }\label{fig:example_tasks}
\end{figure}

See the \href{https://github.com/jazlab/moog.github.io/tree/master/moog_demos/example_configs}{example\_configs} for a variety of task config files.
Four of those are shown in Figure \ref{fig:example_tasks}.
See the \href{https://github.com/jazlab/moog.github.io/tree/master/moog_demos}{demo documentation} for videos of them all and instructions for how to run them with a Python gui.

The MOOG codebase contains libraries of options for each of the components in Section \ref{S:design}, so implementing a task involves only combining the desired ingredients and feeding them to the environment.
For an example, the following code fully implements a navigate-to-goal task, where the subject must move an agent via a joystick action space to a goal location:

\vspace*{-5pt}

\begin{lstlisting}[language=Python]
"""Navigate-to-goal task."""

import collections
from moog import action_spaces, environment, observers, physics, sprite, tasks

# Initial state is a green agent in the center and red goal in the corner
def state_initializer():
    # Goal is a red square with side-length 0.1 at position (0.1, 0.1)
    # Color channels are [c0, c1, c2] arguments
    goal = sprite.Sprite(x=0.1, y=0.1, shape='square', scale=0.1, c0=255)
    # Agent is a green circle at position (0.5, 0.5)
    agent = sprite.Sprite(x=0.5, y=0.5, shape='circle', scale=0.1, c1=255)
    state = collections.OrderedDict([
        ('goal', [goal]),
        ('agent', [agent]),
    ])
    return state

# Physics is a drag force on the agent to limit velocity
phys = physics.Physics((physics.Drag(coeff_friction=0.25), 'agent'))

# Task gives a reward 1 when the agent reaches the goal, and a new trial begins
# 5 timesteps later.
task = tasks.ContactReward(1., 'agent', 'goal', reset_steps_after_contact=5)

# Action space is a joystick with maximum force 0.01 arena widths per timestep^2
action_space = action_spaces.Joystick(0.01, 'agent')

# Observer is an image renderer
obs = observers.PILRenderer(image_size=(256, 256))

# Create the environment, ready to play!
env = environment.Environment(
    state_initializer=state_initializer,
    physics=phys,
    task=task,
    action_space=action_space,
    observers={'image': obs},
}
\end{lstlisting}

\vspace*{-10pt}

This is an extremely simple task, but by complexifying the state initializer and adding additional forces and game rules, a \href{https://github.com/jazlab/moog.github.io/tree/master/moog_demos}{wide range of complex tasks} can be implemented with few lines of code.

\section{Limitations}\label{S:limitations}

Users should be aware of the following limitations of MOOG before choosing to use it for their research:

\begin{itemize}
    \item
    \textbf{Not 3D}.
    MOOG environments are 2.5-dimensional, meaning that they render in 2-dimensions with z-ordering for occlusion.
    MOOG does not support 3D sprites.
    \item
    \textbf{Very simple graphics}.
    MOOG sprites are monochromatic polygons.
    There are no textures, shadows, or other visual effects.
    Composite sprites can be implemented by creating multiple overlapping sprites, but still the graphics complexity is very limited.
    This has the benefit of a small and easily parameterizable set of factors of variation of the sprites, but does make MOOG environments visually unrealistic.
    \item
    \textbf{Imperfect collisions}.
    MOOG's collision module implements Newtonian rotational mechanics, but it is not as robust as professional physics engines (e.g. can be unstable if object are moving very quickly and many collisions occur simultaneously).
\end{itemize}

\section{Related Software}\label{S:related_software}

Professional game engines (e.g. \href{https://unity.com/}{Unity} and \href{https://www.unrealengine.com/}{Unreal}) and visual reinforcement learning platforms (e.g. \href{https://arxiv.org/abs/1612.03801}{DeepMind Lab} \citep{dm_lab}, \href{http://www.mujoco.org/}{Mujoco} \citep{mujoco}, and \href{http://vizdoom.cs.put.edu.pl/}{VizDoom}) are commonly used in the machine learning field for task implementation.
While MOOG has some limitations compared to these (see above), it does also offer some advantages:

\begin{itemize}
    \item
    \textbf{Python}.
    MOOG tasks are written purely in Python, so users who are most comfortable with Python will find MOOG easy to use.
    \item
    \textbf{Procedural Generation}.
    MOOG facilitates procedural generation, with a \href{https://github.com/jazlab/moog.github.io/tree/master/moog/state_initialization/distributions.py}{library of compositional distributions} to randomize conditions across trials.
    \item
    \textbf{Online Simulation}.
    MOOG supports online model-based RL, with a \href{https://github.com/jazlab/moog.github.io/tree/master/moog/env_wrappers/simulation.py}{ground truth simulator} for tree search.
    \item
    \textbf{Psychophysics}.
    MOOG can be run with MWorks, a psychophysics platform.
    \item
    \textbf{Speed}.
    MOOG is fast on CPU.
    While the speed depends on the task and rendering resolution, MOOG typically runs at ~200fps with 512x512 resolution on a CPU, much faster than DeepMind Lab and Mujoco and at least as fast as Unity and Unreal.
\end{itemize}

Python-based physics simulators, such \href{https://pybullet.org/wordpress/}{PyBullet} \citep{pybullet} and \href{http://www.pymunk.org/en/latest/}{Pymunk}, are sometimes used in the psychology literature.
While these offer highly accurate collision simulation, MOOG offers the following advantages:

\begin{itemize}
    \item
    \textbf{Customization}.
    Custom forces and game rules can be easily implemented in MOOG.
    \item
    \textbf{Psychophysics, Procedural Generation, and Online Simulation},
    as described above.
    \item
    \textbf{RL Interface}.
    A task implemented in MOOG can be used out-of-the-box to train RL agents, since MOOG is Python-based and has DeepMind dm\_env and OpenAI Gym interfaces.
\end{itemize}

Psychology and neurophysiology researchers often use platforms such as
\href{https://www.psychopy.org/}{PsychoPy} \citep{psychopy},
\href{http://psychtoolbox.org/}{PsychToolbox} \citep{psychtoolbox}, and
\href{https://mworks.github.io/}{MWorks}.
These allow precise timing control and coordination with eye trackers and other controllers.
MOOG can interface with MWorks to leverage all of those features and offers the following additional benefits:

\begin{itemize}
    \item
    \textbf{Flexibility}.
    MOOG offers a large scope of interactive tasks.
    Existing psychophysics platforms are not easily customized for game-like tasks, action interfaces, and arbitrary object shapes.
    \item
    \textbf{Physics}.
    Existing psychophysics platforms do not have built-in physics, such as forces, collisions, etc.
    \item
    \textbf{RL Interface},
    as described above.
\end{itemize}

\vspace*{0.0in}

\subsubsection*{Acknowledgments}

We thank Chris Stawarz and Erica Chiu for their contributions to the codebase.
We also thank Ruidong Chen, Setayesh Radkani, and Michael Yoo for their feedback as early users of OOG.

\clearpage

\bibliographystyle{abbrvnat}
\setlength{\bibsep}{5pt} 
\setlength{\bibhang}{10pt}

\bibliography{references}

\newpage

\end{document}